\newcommand{\CLASSINPUTinnersidemargin}{18mm}
\newcommand{\CLASSINPUToutersidemargin}{12mm}
\newcommand{\CLASSINPUTtoptextmargin}{20mm}
\newcommand{\CLASSINPUTbottomtextmargin}{25mm}
\documentclass[10pt,conference,a4paper]{IEEEtran}

\usepackage[utf8]{inputenc}

\usepackage{url}

\usepackage{times}

\usepackage{graphicx}
\usepackage{subfigure}
\DeclareGraphicsExtensions{.png,.eps,.ps,.pdf}

\usepackage{url}

\begin{document}

\title{Evaluating Performance of an Adult Pornography Classifier for Child Sexual Abuse Detection}

\author{\IEEEauthorblockN{\small Mhd Wesam Al-Nabki}
\IEEEauthorblockA{\small
Dept. IESA.\\
Universidad de León\\
Researcher at INCIBE\\
mnab@unileon.es}

\and 

\IEEEauthorblockN{\small Eduardo Fidalgo}
\IEEEauthorblockA{\small
Dept. IESA.\\
Universidad de León\\
Researcher at INCIBE\\
eduardo.fidalgo@unileon.es}

\and
\IEEEauthorblockN{\small Roberto A. Vasco-Carofilis}
\IEEEauthorblockA{\small
Dept. IESA.\\
Universidad de León\\
Researcher at INCIBE\\
rvasc@unileon.es}

\and 

\IEEEauthorblockN{\small Francisco Jañez-Martino}
\IEEEauthorblockA{\small
Dept. IESA.\\
Universidad de León\\
Researcher at INCIBE\\
fjanm@unileon.es}

\and
\IEEEauthorblockN{\small Javier Velasco-Mata}
\IEEEauthorblockA{\small
Dept. IESA.\\
Universidad de León\\
Researcher at INCIBE\\
jvelm@unileon.es
}}

\maketitle

\begin{abstract}
The information technology revolution has facilitated reaching pornographic material for everyone, including minors who are the most vulnerable in case they were abused. Accuracy and time performance are features desired by forensic tools oriented to child sexual abuse detection, whose main components may rely on image or video classifiers.
In this paper, we identify which are the hardware and software requirements that may affect the performance of a forensic tool. We evaluated the adult porn classifier proposed by Yahoo, based on Deep Learning, into two different OS and four Hardware configurations, with two and four different CPU and GPU, respectively. The classification speed on Ubuntu Operating System is $~5$ and $~2$ times faster than on Windows 10, when a CPU and GPU are used, respectively. We demonstrate the superiority of a GPU-based machine rather than a CPU-based one, being $7$ to $8$ times faster. Finally, we prove that the upward and downward interpolation process conducted while resizing the input images do not influence the performance of the selected prediction model. 

\end{abstract}


{\bf Type of contribution:}  {\it Short original research}

\section{Introduction}

Possession of Child Sexual Abuse Material (CSAM) is one of the most terrible crimes against children because it involves the sexual and violent abuse of innocent minors. Manual search for evidence in a seized hard drive can be a long and complex process due to the enormous number of files. Furthermore, when it comes to finding illegal material in the field of the police search, reliability and speed are essential. This is because the police forces have a limited time to search CSAM content on seized devices, and within this time slot, they can differentiate between taking the suspect in detention or not.

\begin{figure}[ht]
\centering
\includegraphics[width=\columnwidth]{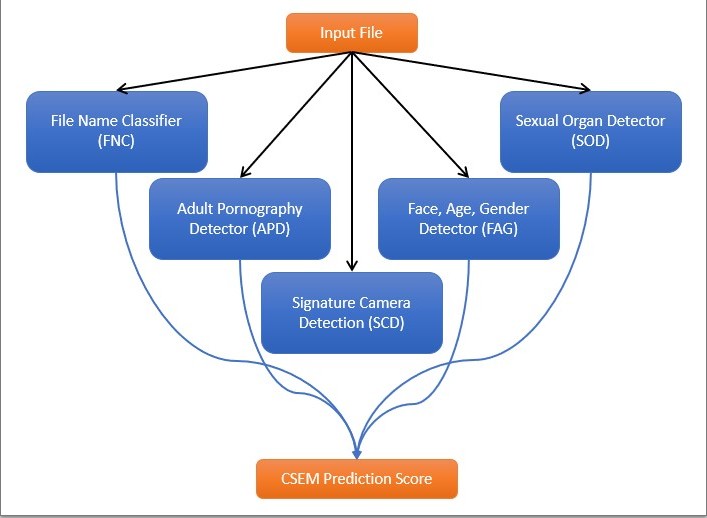}
\caption{An overview of the 4NSEEK modules that are involved in identifying CSAM. The framework receives an input file that is analyzed by FNC, FAG, APD, and SOD. Based on each module output, a CSAM Prediction Score is built, which ranges from $0$ to $9$, and it represents a probability of being Safe (closer to $0$) or CSAM (closer to $9$).}
\label{fig:4NSEEK}
\end{figure}

This paper is part of the European project Forensic Against Sexual Exploitation of Children (4NSEEK) \cite{incibe_2020}, and its primary goal is to provide a forensic tool to detect CSAM via the combination of several modules: File Name Classifier (FNC) \cite{fnc_icpram20}, Sexual Organ Detector (SOD) \cite{sod_icpram20}, Signature Camera Detection (SCD) \cite{scd_icpram20}, Adult Pornography Detector (APD) \cite{gangwar2017pornography} and a Face detector, Age and Gender (FAG) estimator \cite{chaves2019improving, ICDP2019Chaves, desiy_visapp20}. All these systems work simultaneously to identify CSAM (Fig. \ref{fig:4NSEEK}).

The speed and confidence of the prediction are critical when investigating a crime related to child sexual abuse.
This paper focuses on finding the optimum hardware and software requirements to obtain the best performance of the APD system. Specifically, this paper attempts to answer three crucial questions: 1) what is the best Operating System (OS) to be used for deploying the software, Windows or Linux OS?, 2) what is the prediction speed using a Graphical Processing Unit (GPU) and a Central Processing Unit (CPU)? and 3) does the resizing of the input image, using an upward or a downward interpolation function, affects the performance of the classifier in terms of accuracy and processing time? 

The rest of the paper is organized as follows: section \ref{sec: state-of-the-art} presents the related work. Then, section \ref{sec: methodology} introduces the used neural network model. After that, in section \ref{sec:empirical_Settings}, we present the used dataset for the conducted experiments as well as the hardware and the software specifications of the used computer machines. Next, section \ref{sec: result} demonstrates the obtained results. Finally, Section \ref{sec:conclusions_and_future_work} presents our conclusions and points out to our future work.

\begin{figure*}[ht]
\centering
\includegraphics[width=\textwidth]{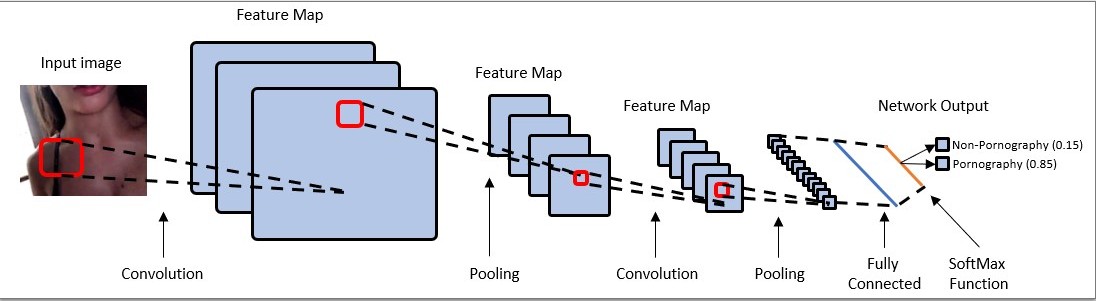}
\caption{A graphical representation of the Adult Pornography Detection model}
\label{fig:apd}
\end{figure*}

\section{Literature Review}
\label{sec: state-of-the-art}
Several researchers have addressed the problem of identifying pornography images. A traditional strategy to identify nudity in images depends on detecting human skin in the image using color \cite{lin2003pornography, zuo2010patch} and/or texture \cite{sathish2008texture}. When an input image contains a high percentage of pixels with colors close to the skin, it is considered as an indicator of nudity. However, this signal solely is not reliable since a face and hands images have many skin pixels while being non-porn. Also, the color of the skin has a wide range that can match with other objects in the input image. To cope with this limitation, researchers have developed a bag of visual words (BOVW) model that attempts to extract the most frequent patches that exist on a set of training images and try to find it in the test images~\cite{deselaers2008bag, ries2014survey}.

The rise of Deep Learning (DL) techniques through the automatic feature extraction have revolutionized the state-of-the-art performance \cite{moustafa2015applying, wang2016pornographic, perez2017video, gangwar2017pornography, wehrmann2018adult, Islam2019, chaves2019improving, yahoo_2020, austin2020classifying}.
Yahoo Inc. proposed Not Suitable for Work (NSFW) convolutional neural network model \cite{yahoo_2020} to identify adult pornography images. Moustafa et al. \cite{moustafa2015applying} used a combination of ConvNets, whereas they fused and fine-tuned AlexNet and GoogLeNet to adapt these models to pornographic data. Their model has shown a remarkable increase in the classification accuracy on the NPDIP Pornographic-$800$ and Pornographic-$2$k datasets \cite{avila2013pooling}. Wang et al. presented a novel approach, called Strongly-supervised Deep Multiple Instance Learning (SDMIL) that models each input image as a bag of overlapped image patches, and they trained the model as a Multiple Instance Learning problems. Wehrmann et al. \cite{wehrmann2018adult} used a Convolutional Neural Network and long short-term memory (LSTM) recurrent networks for detecting pornography content.

\section{Methodology}
\label{sec: methodology}
To build the Adult Pornography Detector (APD), we adopted the Not Suitable for Work (NSFW) \cite{mahadeokar_pesavento_2016} model because it is dedicated to recognizing pornography images. A graphical representation of the model is shown in Figure \ref{fig:apd}.
The NSFW model uses ResNet-$50$-thin architecture as a pre-trained network \cite{he2016deep}, which was trained on $1,000$ ImageNet dataset classes~\cite{deng2009imagenet}. To adapt the ResNet-$50$-thin to a binary classifier, only the last layer was replaced with a two nodes fully-connected layer. After that, the weights of the model were find-tuned on the NSFW dataset. 
Since the NSFW image classification model expects an input image size to be $256X256$ pixels, a pre-processing function is called to resize the image to the desired size before predicting its category. Two popular techniques were proposed to change the size of the input image to fit with the input size of the model \cite{hashemi2019enlarging}; they are padding with zeros or interpolation. In this work, the APD module adopts the latter approach to resize the input image into the desired size.
 
\section{Experimental Settings}
\label{sec:empirical_Settings}
\begin{figure}[htp]
\subfigure[Pornography class]{\includegraphics[width=0.23\textwidth]{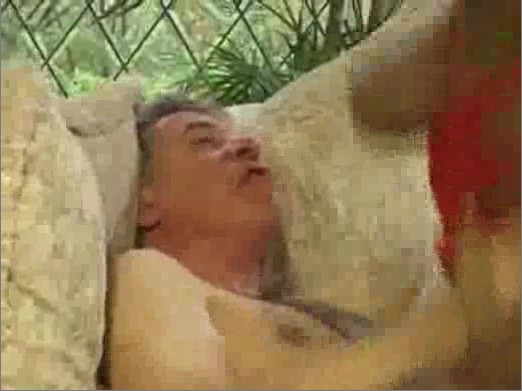}}
\hspace*{\fill}
\subfigure[Non-pornography class]{\includegraphics[width=0.23\textwidth]{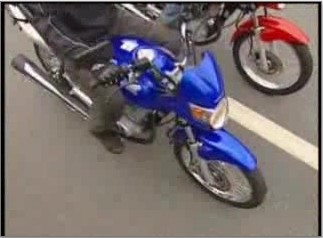}}
\caption{Samples from the Pornography Database}
\label{fig:apd_samples}
\end{figure}

To measure the performance of the APD module, We proposed a test set of $6,000$ images, randomly selected from the Pornography Database\footnote{\url{https://sites.google.com/site/pornographydatabase/}} \cite{avila2013pooling}. The dataset is balanced whereas the non-pornographic and the pornographic classes have the same number of samples, i.e. $3,000$ images. Fig. \ref{fig:apd_samples} shows samples of both categories of the dataset. 

\begin{figure}[htp]
\subfigure{\includegraphics[width=0.23\textwidth]{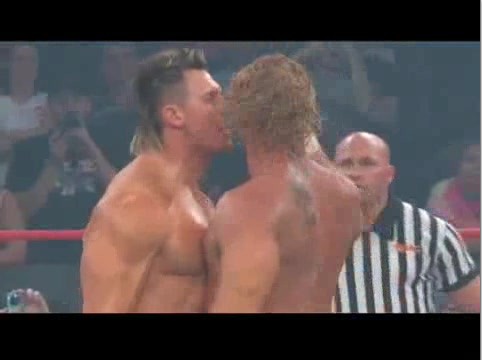}}
\hspace*{\fill}
\subfigure{\includegraphics[width=0.23\textwidth] {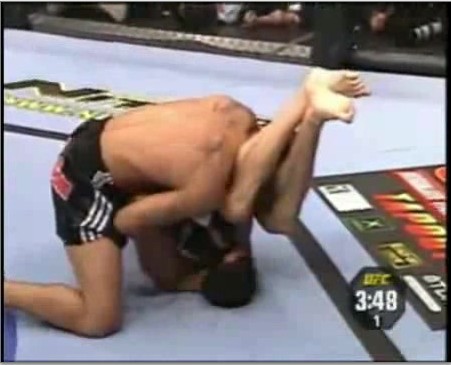}} 
\caption{Challenging samples from the non-pornography class}
\label{fig:apd_samples_neg}
\end{figure}

It can be observed that the dataset contains challenging images that expose skin explicitly, while they are not pornographic, such as the samples illustrated in Fig. \ref{fig:apd_samples_neg}. 

\begin{figure}[htp]
\subfigure{\includegraphics[width=0.23\textwidth]{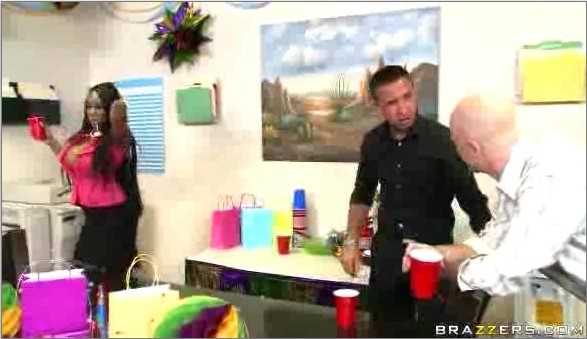}}
\hspace*{\fill}
\subfigure{\includegraphics[width=0.23\textwidth] {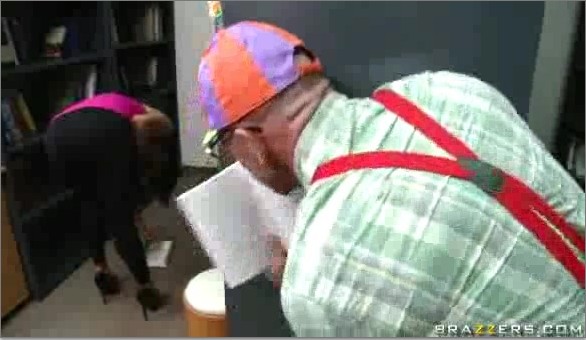}} 
\caption{Challenging samples from the pornography class}
\label{fig:apd_samples_pos}
\end{figure}

In contrast, other images do not involve skin exposure but they refer to the pornography class, like the samples shown in Fig. \ref{fig:apd_samples_pos}.

Table \ref{tab:machines_specifications} presents the used computer machines to conduct the experiments of this paper. All the used machines are provided with Ubuntu $18.04$ OS, except machine \#$4$, which has a dual boot OS of Windows 10 and Ubuntu.

\begin{table}[htp]
\centering
\caption{Specifications of the computer machines used to evaluated the APD performance. The letter \textit{M.\#} stands for machine.}
\label{tab:machines_specifications}

\begin{tabular}{c c c}
\hline
\hline
Machine ID &
\multicolumn{1}{c}{\begin{tabular}[c]{@{}c@{}}GPU Model/ \\ Memory (GB)\end{tabular}} &
\multicolumn{1}{c}{\begin{tabular}[c]{@{}c@{}}CPU Model/ \\ Memory (GB)\end{tabular}} 

\\  \hline

\multicolumn{1}{c}{M. 1} &
\begin{tabular}[c]{@{}l@{}}Nvidia RTX 2060/\\6GB GDDR6\end{tabular} &
\begin{tabular}[c]{@{}l@{}}Intel Core i7/\\16GB\end{tabular} 
\\
\multicolumn{1}{c}{M. 2} &
\begin{tabular}[c]{@{}l@{}}Nvidia RTX 2070/\\8 GB GDDR6\end{tabular} &     
\begin{tabular}[c]{@{}l@{}}Intel Core i7/\\8GB\end{tabular}     
\\
\multicolumn{1}{c}{M. 3} &
\begin{tabular}[c]{@{}l@{}}Nvidia GTX 1050/\\4 GB GDDR5\end{tabular} &
\begin{tabular}[c]{@{}l@{}}Intel Core i9/\\32GB\end{tabular}
\\
\multicolumn{1}{c}{M. 4} & 
\begin{tabular}[c]{@{}l@{}}Nvidia GTX 1060/\\6 GB GDDR5\end{tabular} & 
\begin{tabular}[c]{@{}l@{}}Intel Core i7/\\16GB\end{tabular} 
\\

\hline
\hline

\end{tabular}

\end{table}

\section{Experimental Results}
\label{sec: result}

\subsection{Operating System Selection}
To answer the first research question raised in this paper concerning the selection of the operating system, we evaluated the prediction speed on machine \#$4$ (Table \ref{tab:machines_specifications}), which hosts two operating systems. 
We found that the sequential prediction of the test set samples took $1,298.59$s and $284.79$s using the CPU on the Windows and Ubuntu machines, respectively.
Using the GPU of machine \#$4$, we observed similar behavior, whereas the prediction speed on the Windows machine was slower than the Ubuntu one with $178.8$s and $88.19$s, respectively. Hence, Ubuntu OS is, at least, $5$ and $2$ times faster than Windows in CPU and GPU, respectively. This behavior could be due to the high number of processes running the background in Windows OS in comparison to Linux-based OS \cite{misal_2020}. Therefore, we conclude that regardless of the back-end hardware, the operating system has a notable impact on the prediction speed. Hence, based on our analysis, we would recommend building the APD over an Ubuntu OS. 

\subsection{Processing Unit Selection}
The second research question addressed in this paper attempts to estimate the time needed to predict the samples of the test set over several GPUs and CPUs machines. However, given the superiority of Ubuntu OS, hereafter, it is used for the next experiments. The specification of the examined machines and the consumed time are presented in Table \ref{tab:machines_processing_time}. Our results indicate that using a GPU-based machine is always faster than using a CPU-based machine. Also, Table \ref{tab:machines_processing_time} shows that machine \#$1$, which uses Nvidia RTX $2060$, is the best graphical card among the benchmarked ones. Concerning the CPU machines, we observed that machine \#$3$, which operates on Intel Core i$9$, is the best CPU for this task in comparison to the explored processors. 

\begin{table}[htp]
\centering
\caption{The processing time on several GPU and CPU machines. The values in bold font refer to the fastest CPU and GPU machines.}
\label{tab:machines_processing_time}

\begin{tabular}{c c c}
\hline
\hline
Machine ID &
\multicolumn{1}{c}{\begin{tabular}[c]{@{}c@{}} GPU Processing\\Time (seconds)\end{tabular}} &
\multicolumn{1}{c}{\begin{tabular}[c]{@{}c@{}}CPU Processing\\Time (seconds)\end{tabular}}
\\ \hline
M. 1 & \textbf{57.88} & 589.13 \\
M. 2 & 80.61 & 493.25 \\
M. 3 & 86.61 & \textbf{442.61} \\
M. 4 & 89.19 & 453.43 \\
\hline
\hline

\end{tabular}

\end{table}

\subsection{Upward/Downward Image Resizing Impact}
Lastly, we analyze the impact of resizing the input image, either upward or downward, on the speed and the accuracy of the prediction. The APD module expects an image of $256X256$ pixels. However, in the real case scenario, the size of the input image may vary significantly, as it might be smaller or larger than the desired size. Typically, a pre-processing function is called to resize them upward or downward. In this experiment, we downscale the input images by $25\%$, $50\%$, $75\%$, and $100\%$ (the latter size refers to the original input size of the image, without resizing). Next, to feed the APD module with the input image, we call the pre-processing function to adjust the image size to the correct input size, i.e., $256X256$ pixels. 

Table \ref{tab:performance_resize_time} shows that resizing the input image does not affect the prediction time adversely. Instead, we observed faster performance when the images were downscaled before feeding it to the APD model. In our experiments, we realized that resizing the input images into $25\%$ of their original size obtained the fastest prediction time.

\begin{table}[htp]
\centering
\caption{The performance of the image classifier on the test set in terms of time. The values in bold font refer to the best accuracy obtained.}
\label{tab:performance_resize_time}

\begin{tabular}{ccc}
\hline
\hline
Resize (\%) & \begin{tabular}[c]{@{}c@{}}Nvidia RTX 2060\\  (seconds)\end{tabular} & \begin{tabular}[c]{@{}c@{}}Intel Core i9\\ (seconds)\end{tabular} \\ \hline
100\% & 57.88 & 442.61 \\
75\% & 50.17 & 435.95 \\
50\% & 48.32 & 439.68 \\
\textbf{25\%} & \textbf{47.38} & \textbf{428.42}
\\
\hline
\hline
\end{tabular}

\end{table}

Additionally, we estimated the accuracy of the APD model after resizing the images, as shown in table \ref{tab:performance_resize_accuracy}. Interestingly, we did not record significant changes in the prediction performance of the model using the other resize values did not influence the prediction accuracy, except when resizing the image to $25\%$ of its original size. In this case, the F1 score of the model increased from $0.73$ to $0.74$.
Therefore, we can conclude that this upward and downward interpolation process to adjust the input image size does not affect the performance negatively, and it may lead to a positive impact.

\begin{table}[htp]
\centering
\caption{The performance of the image classifier on the test set in terms of accuracy and F1 score. The values in bold font refer to the best accuracy obtained.}
\label{tab:performance_resize_accuracy}

\begin{tabular}{ccccc}
\hline
\hline
Resize (\%) & \multicolumn{1}{c}{Precision} & \multicolumn{1}{c}{Recall} & \multicolumn{1}{c}{F1 Score} & \multicolumn{1}{c}{Accuracy} \\ \hline
\multicolumn{1}{c}{100\%} & 0.78 & 0.74 & 0.73 & 0.74 \\
\multicolumn{1}{c}{75\%} & 0.78 & 0.74 & 0.73 & 0.74 \\
\multicolumn{1}{c}{50\%} & 0.78 & 0.74 & 0.73 & 0.74 \\
\multicolumn{1}{c}{\textbf{25\%}} & \textbf{0.81} & \textbf{0.75} & \textbf{0.74} & \textbf{0.75}\\
\hline
\hline
\end{tabular}

\end{table}

\section{Conclusion and Future Work}
\label{sec:conclusions_and_future_work}
This paper analyzed the performance of Adult Pornography Detector (APD), which is a core component of the Forensic Against Sexual Exploitation of Children (4NSEEK) project to identify Child Sexual Abuse Material (CSAM). The APD adopted the Not Suitable for Work (NSFW) model to detect pornography images, and we established our experimentation on a balanced dataset of $6,000$ images selected randomly from the Pornography Database.

Our analysis discovered that deploying the APD on an Ubuntu OS is faster than Windows 10 in terms of prediction time. Ubuntu OS was, at least, $5$ and $2$ times faster than Windows 10 in CPU and GPU machines, respectively. 
Furthermore, we found that using a GPU-based machine, i.e. Nvidia RTX 2060, is $7$ to $8$ times faster than a CPU-based machine, i.e. Intel Core i$9$, with a processing time of $57.88$s and $442.61$s, respectively.
Finally, we realized that APD is robust against the upward and downward resizing of the input image on the classifier's accuracy and speed. Also, we observed a slight improvement in the prediction accuracy and the processing time when the input images were downscaled to $25\%$ of its original size.

In the future, we plan to enhance the performance of the base classification model. Concretely, we want to explore advanced pre-trained models, such as Inception Resnet \cite{szegedy2017inception} and MobileNetV2 \cite{sandler2018mobilenetv2}.

\section*{Acknowledgements}

This work was supported by the framework agreement between the University of Le\'on and INCIBE (Spanish National Cybersecurity Institute) under Addendum 01. This research has been funded with support from the European Commission under the 4NSEEK project with Grant Agreement 821966. This publication reflects the views only of the author, and the European Commission cannot be held responsible for any use which may be made of the information contained therein. We acknowledge NVIDIA Corporation with the donation of the TITAN Xp and Tesla K40 GPUs used for this research.

\end{document}